\documentclass[a4paper, 10 pt, conference]{hsmr} 
\usepackage[utf8]{inputenc}
\IEEEoverridecommandlockouts                       
\usepackage{mathtools}
\usepackage{geometry}
\usepackage[labelsep=space]{caption}
\usepackage{newtxmath,newtxtext}
\usepackage{authblk}
\usepackage{pgfplots}
\usepackage{graphicx}
\usepackage[colorlinks,urlcolor=blue]{hyperref} 
\usepackage{newtxtext,newtxmath}
\usepackage{booktabs}
\usepackage{multirow}
\usepackage{multicol}
\usepackage{textcomp}
\usepackage{tabularx}
\usepackage{array}
\pgfplotsset{compat=newest} 

\usepackage{fancyhdr}

\fancypagestyle{firstpage}{
    \fancyhf{}
    \fancyhead[CO]{\color{blue}HSMR2024: THE 16TH HAMLYN SYMPOSIUM ON MEDICAL ROBOTICS}
}

\title{\LARGE \bf
UltraGelBot: Autonomous Gel Dispenser for Robotic Ultrasound\vspace{-0.4cm}} 
\author{\LARGE Deepak Raina$^{1,2}$, Ziming Zhao$^{1}$, Richard Voyles$^{1}$, Juan Wachs$^{1}$,\\ Subir K. Saha$^{2}$,  S. H. Chandrashekhara$^{3}$ \vspace{-0.4cm}}
\affil{\Large\textit{ $^{1}$Purdue University, $^{2}$Indian Institute of Technology Delhi, $^{3}$AIIMS Delhi}\vspace{-0.35cm}}
\affil{\Large\textit{draina@purdue.edu}\vspace{-0.35cm}}
\begin{document}

\maketitle 
\thispagestyle{firstpage}

\section*{INTRODUCTION}
\vspace{-0.15cm} 
Robotic Ultrasound Systems (RUSs) have gained increasing prominence in recent years to alleviate problems concerning operator-dependability in free-hand ultrasound examinations \cite{jiang2023robotic}. These robotic systems operate either in teleoperated \cite{raina2021comprehensive} or autonomous \cite{raina2023robotic, raina2023rusopt, raina2023deep} mode. These systems can increase ultrasound accessibility in underserved regions and ensure the safety of the operator during the COVID-19 pandemic \cite{raina2021comprehensive, al2021autonomous}. However, in these robotic systems, the gel is still applied by the human attendant. This human intervention between the procedure results in the acquisition of sub-optimal images due to inappropriate acoustic coupling \cite{wang2021continuous}. Further, the procedure time also increases as RUS needs to be halted several times in between the procedure for manual gel application. Moreover, the presence of humans near the patient's vicinity did not achieve the complete safety of the operator, as promised by RUS. Thus, an automated method for dispensing of ultrasound gel is the immediate requirement of RUS. 

In this paper, we proposed a new end-of-arm tool for RUS, referred to as `UltraGelBot'. This robot can detect and dispense the gel autonomously. A deep learning model is developed to detect the gel region from real-time images acquired using an onboard camera. A motorized mechanism is developed, which uses this feedback and dispenses the gel. In addition, sonographers often utilize multiple ultrasound probes, such as linear, curvilinear, etc., to visualize different anatomies up to a specific depth. The existing gripping methods for RUS are specific to the design profiles of the probe. Thus, this gripper features a modular mechanism to grip ultrasound probes with varying profiles. This mechanism will also allow the operator to easily change different probes during the procedure without using mechanical tools. The performance of the UltraGelBot was compared to that of manual gel application for ultrasound of artery in human's forearm. In addition, for assessment of user experience, the NASA Task Load Index
(TLX) was used.
\section*{MATERIALS AND METHODS}
\vspace{-0.1cm}
\subsection{Design}
\begin{figure}[t]
	\centering
	%trim={L,B,R,T}
	\includegraphics[trim=0cm 3cm 0cm 0cm,clip,width=\linewidth]{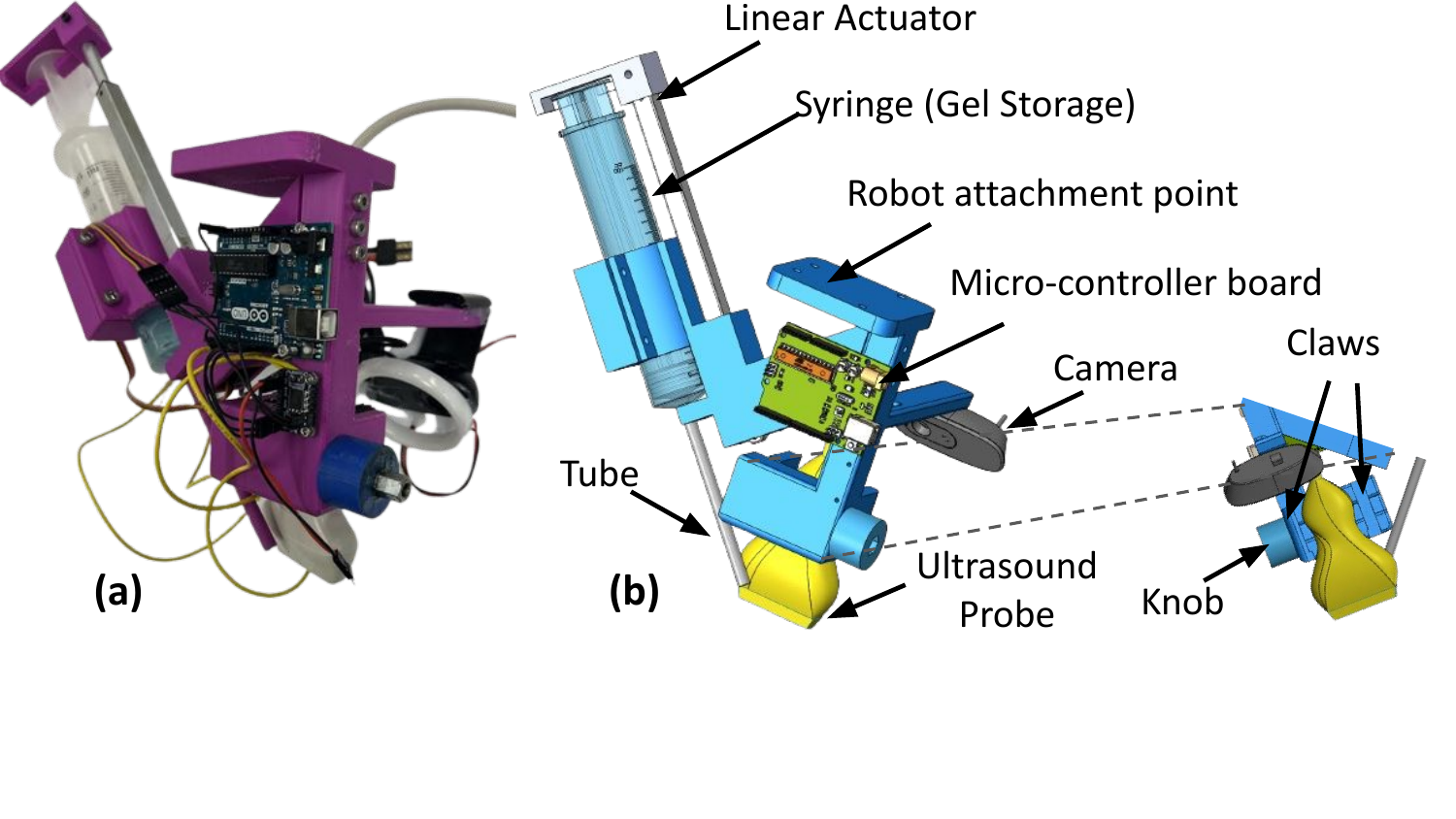}
	\caption{(a) Prototype (b) Isometric view of the CAD model highlighting the components of the UltraGelBot \vspace{-0.1cm}}
	\label{fig:design}
\end{figure}

% \vspace{-0.2cm}
Fig. \ref{fig:design}(a) and \ref{fig:design}(b) show the isometric view of the prototype and  3D CAD model of the UltraGelBot, respectively. 
% The design has two main parts: the probe gripper and the gel dispenser. 
The design features a solid tube pointing at the side of the probe through which the gel was applied to the body surface. This tube is connected to a reservoir, which is a syringe for this prototype. The syringe is constrained by the housing to stabilize its linear movement. The syringe has a piston, which is connected to the moving end of the linear actuator (Actuonix L16-P). It will move to compress the piston inside the reservoir and dispense the gel. For refilling, the syringe needs to be detached from the assembly and refilled using a commercially used container. Larger reservoirs may be included in the design in order to reduce the frequency of refilling during the procedure. 

The ultrasound probe will be gripped between two claws, one of which is movable. The knob-screw mechanism has been used to allow for the closing of the claws. Interestingly, it can be operated without requiring any mechanical tools, which allows the operator to easily change the probe if required. Further, the adjustable claw of this design allows it to accommodate several commercial probes with different design profiles.

\vspace{-0.08cm}
\subsection{Gel Detection}
\begin{figure}[ht]
	\centering
	%trim={L,B,R,T}
	\includegraphics[width=\linewidth, trim={0cm 9cm 1cm 0cm}, clip]{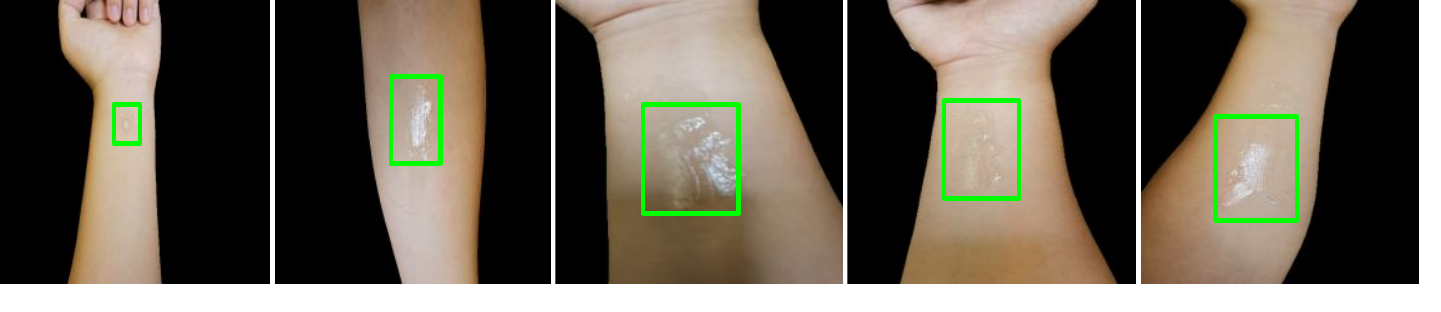}
	\caption{Dataset used for training of gel detection model}
	\label{fig:gel_dataset}
\end{figure}

The design included a method to detect the existence of ultrasound gel between the ultrasound probe and the human body. The detection was performed using a single camera by analyzing the region of the body in contact with the probe. The gel detection algorithm implemented in this design was based on the Faster Region-based Convolutional Neural Network (F-RCNN) \cite{ren2015faster}, which is a state-of-the-art object detection model. To train and test the algorithm, a labeled dataset was created, as illustrated in Fig. \ref{fig:gel_dataset}. The dataset consists of images of people with different skin conditions, each containing varying amounts of ultrasound gel applied to the skin. The dataset was annotated with rectangular entities that represent the region of interest (RoI), i.e., gel on the surface of the human's body.
% \subsubsection{\color{black}Gel Dataset}
\subsection{Gel Dispenser}
During runtime, the model detected multiple RoIs with confidence values associated with each one. The RoI with the highest confidence was considered as the region where the ultrasound gel was detected. These regions were used to detect if the movement of the ultrasound probe left a trail of gel on the human's body. The existence of the gel trail is determined by checking if it exists in the region where the probe has just passed. If the gel was not detected, the additional gel would be supplied by activating the linear actuator. The dispenser is activated through a logic written on the Arduino microcontroller, which is connected to the GPU system. The gripper camera is also connected to the system. The Python script was written to create an interface between the microcontroller logic, dispenser unit, detection model, and camera feedback. This script activates the dispenser unit when it receives feedback from the detection model regarding the absence of gel. 
\section*{RESULTS}
\begin{figure}[!htbp]
	%trim={L,B,R,T}
	\centering
	\includegraphics[width=\linewidth, trim={0cm 8.6cm 0.6cm 0cm},clip]{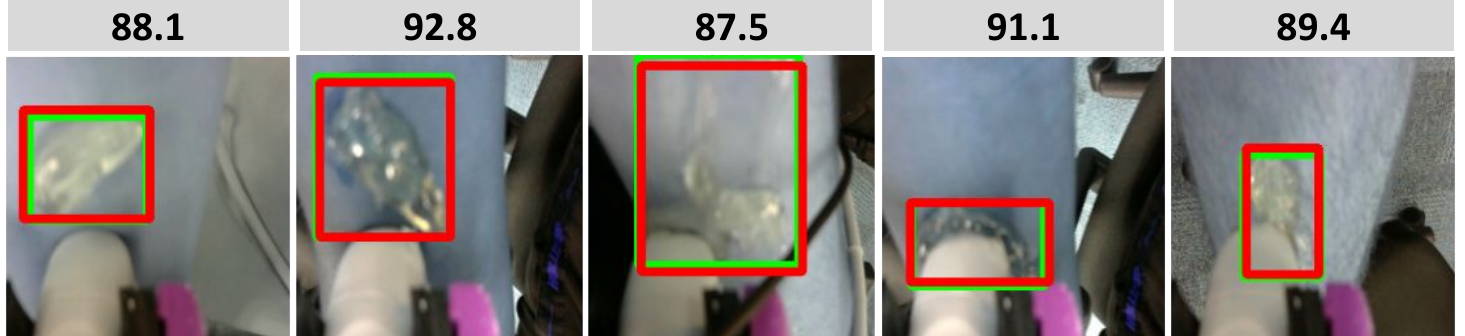}
	\caption{Ground truth (green) and predicted (red) RoI by the gel detection model along with their IoU values.}
	\label{fig:gel_detection}
\end{figure}

\vspace{-0.1cm}
The gel detection model was validated on a test dataset, and results reported a Mean Intersection over Union (IoU) of $0.87$. The model's performance was further improved by fine-tuning it on a smaller dataset of gel images captured in real-time by the system, resulting in a mean IoU of $0.91$. In addition to the quantitative metrics, the qualitative inspection of the model's output on human data is shown in Fig. \ref{fig:gel_detection}. This demonstrated that the algorithm was able to accurately localize the gel on the human skin with probe movement. 

% \vspace{-0.3cm}
Table \ref{tab:performance} compares the performance of the ultrasound scanning of artery in the human's forearm with gel applied using UltraGelBot to the manual application. The performance metrics include mean image quality and scanning time. The image quality has been computed using the deep learning model from our previous work \cite{raina2024deep}. The study involved 3 participants. Each participant conducted the procedure 6 times, 3 each with gel application by manual and UltraGelBot, respectively. The manual group showed a mean image quality of $2.58$, while the UltraGelBot group showed $3.06$ ($18.6\%$ increase). In addition, the scanning time also decreased from $14.5$ seconds in the manual group to $9.1$ seconds ($37.2\%$ decrease) in the UltraGelBot group. The results of NASA TLX were statistically analyzed using standard questionnaires after each experiment. The participants experienced significantly less mental, physical, and temporal demands, less effort and frustration, and achieved better performance with the UltraGelBot.

\setlength{\tabcolsep}{2pt}
\renewcommand{\arraystretch}{1.1}
{\begin{table}[!ht]
    \centering
\caption{Comparison of manual gel application to UltraGelBot. M: Mental, P: Physical, T: Temporal, P*: Performance, E: Effort, F: Frustration}
\vspace{-1mm}
\resizebox{\linewidth}{!}{\begin{tabular}{ccccccccc}  
    % {\begin{tabular}{ccccccc}  
    \toprule
    \textbf{Ultrasound} & \multicolumn{2}{c}{\textbf{Performance metrics}} & \multicolumn{6}{c}{\textbf{NASA TLX weight metrics}} \\
    \cmidrule(lr){2-3}
    \cmidrule(lr){4-9}
    \textbf{Gel} &  Mean image &  Scanning & \multicolumn{3}{c}{Demand} &  \multirow{2}{*}{P*} &  \multirow{2}{*}{E} &
    \multirow{2}{*}{F} \\
    \cmidrule(lr){4-6}
    \textbf{Source} & quality & time (sec.) & M & P & T & ~ & ~ & ~ \\
    \midrule
    \textbf{Manual} & $2.58\pm0.7$ & $14.5\pm3.0$ & 7.7 & 6.7 & 7.6 & 6 & 5.7 & 7.3\\
    \textbf{UltraGelBot} & $3.06\pm1.1$ & $9.1\pm1.7$ &  5.3 & 5 & 4.7 & 9.0 & 4.0 & 4.0 \\
    \midrule
    \textbf{Change} & $\uparrow18.6\%$ & $\downarrow37.2\%$ & $\downarrow30\%$ & $\downarrow25\%$ & $\downarrow39\%$ & $\uparrow50\%$ & $\downarrow29\%$ & $\downarrow45\%$\\
    \bottomrule
    \end{tabular}}
    \label{tab:performance}
\end{table}}

\vspace{-0.2cm}
\section*{DISCUSSION}
\vspace{-0.1cm}
In this study, an autonomous gel dispenser was designed for robotic ultrasound systems. An onboard camera was used to capture the images, and a deep learning model was developed to detect the gel from images. This dispenser improved the quality of the images acquired, reduced scanning time and was more effective from the users’ perspective. 
One major limitation of this prototype is that the movement of probe is constrained in one direction due to single camera. 
In future designs, use of multiple cameras and optical trackers will be explored to enable probe movement in multiple directions. Further, its applicability will be tested for \textit{in-vivo} clinical procedures.  

\bibliographystyle{IEEEtran}
\bibliography{HSMR}

\begin{thebibliography}{1}
\providecommand{\url}[1]{#1}
\csname url@samestyle\endcsname
\providecommand{\newblock}{\relax}
\providecommand{\bibinfo}[2]{#2}
\providecommand{\BIBentrySTDinterwordspacing}{\spaceskip=0pt\relax}
\providecommand{\BIBentryALTinterwordstretchfactor}{4}
\providecommand{\BIBentryALTinterwordspacing}{\spaceskip=\fontdimen2\font plus
\BIBentryALTinterwordstretchfactor\fontdimen3\font minus \fontdimen4\font\relax}
\providecommand{\BIBforeignlanguage}[2]{{%
\expandafter\ifx\csname l@#1\endcsname\relax
\typeout{** WARNING: IEEEtran.bst: No hyphenation pattern has been}%
\typeout{** loaded for the language `#1'. Using the pattern for}%
\typeout{** the default language instead.}%
\else
\language=\csname l@#1\endcsname
\fi
#2}}
\providecommand{\BIBdecl}{\relax}
\BIBdecl

\bibitem{jiang2023robotic}
Z.~Jiang, S.~E. Salcudean, and N.~Navab, ``Robotic ultrasound imaging: State-of-the-art and future perspectives,'' \emph{Medical image analysis}, p. 102878, 2023.

\bibitem{raina2021comprehensive}
D.~Raina, H.~Singh, S.~K. Saha, C.~Arora, A.~Agarwal, S.~Chandrashekhara, K.~Rangarajan, and S.~Nandi, ``Comprehensive telerobotic ultrasound system for abdominal imaging: Development and in-vivo feasibility study,'' in \emph{2021 International Symposium on Medical Robotics (ISMR)}.\hskip 1em plus 0.5em minus 0.4em\relax IEEE, 2021, pp. 1--7.

\bibitem{raina2023robotic}
D.~Raina, S.~Chandrashekhara, R.~Voyles, J.~Wachs, and S.~K. Saha, ``Robotic sonographer: Autonomous robotic ultrasound using domain expertise in bayesian optimization,'' in \emph{2023 IEEE International Conference on Robotics and Automation (ICRA)}.\hskip 1em plus 0.5em minus 0.4em\relax IEEE, 2023, pp. 6909--6915.

\bibitem{raina2023rusopt}
D.~Raina, A.~Mathur, R.~M. Voyles, J.~Wachs, S.~Chandrashekhara, and S.~K. Saha, ``Rusopt: Robotic ultrasound probe normalization with bayesian optimization for in-plane and out-plane scanning,'' in \emph{2023 IEEE 19th International Conference on Automation Science and Engineering (CASE)}.\hskip 1em plus 0.5em minus 0.4em\relax IEEE, 2023, pp. 1--7.

\bibitem{raina2023deep}
D.~Raina, S.~Chandrashekhara, R.~Voyles, J.~Wachs, and S.~K. Saha, ``Deep kernel and image quality estimators for optimizing robotic ultrasound controller using bayesian optimization,'' in \emph{2023 International Symposium on Medical Robotics (ISMR)}.\hskip 1em plus 0.5em minus 0.4em\relax IEEE, 2023, pp. 1--7.

\bibitem{al2021autonomous}
L.~Al-Zogbi, V.~Singh, B.~Teixeira, A.~Ahuja, P.~S. Bagherzadeh, A.~Kapoor, H.~Saeidi, T.~Fleiter, and A.~Krieger, ``Autonomous robotic point-of-care ultrasound imaging for monitoring of covid-19--induced pulmonary diseases,'' \emph{Frontiers in Robotics and AI}, vol.~8, p. 645756, 2021.

\bibitem{wang2021continuous}
C.~Wang, B.~Qi, M.~Lin, Z.~Zhang, M.~Makihata, B.~Liu, S.~Zhou, Y.-h. Huang, H.~Hu, Y.~Gu \emph{et~al.}, ``Continuous monitoring of deep-tissue haemodynamics with stretchable ultrasonic phased arrays,'' \emph{Nature biomedical engineering}, vol.~5, no.~7, pp. 749--758, 2021.

\bibitem{ren2015faster}
S.~Ren, K.~He, R.~Girshick, and J.~Sun, ``Faster r-cnn: Towards real-time object detection with region proposal networks,'' \emph{Advances in neural information processing systems}, vol.~28, 2015.

\bibitem{raina2024deep}
D.~Raina, S.~H. Chandrashekhara, R.~Voyles, J.~Wachs, and S.~K. Saha, ``Deep learning model for quality assessment of urinary bladder ultrasound images using multi-scale and higher-order processing,'' \emph{IEEE Transactions on Ultrasonics, Ferroelectrics, and Frequency Control}, pp. 1--1, 2024.

\end{thebibliography}

\end{document}